\documentclass[11pt]{article}

\usepackage[final]{acl}

\usepackage{times}
\usepackage{latexsym}

\usepackage[T1]{fontenc}

\usepackage[utf8]{inputenc}

\usepackage{microtype}

\usepackage{inconsolata}

\usepackage{amsmath,nccmath,tabularx} 
\usepackage{algorithm}
\usepackage{algorithmic}
\usepackage{color}
\usepackage{graphicx}
\graphicspath{{images/}}
\usepackage{wrapfig,lipsum}
\usepackage{lscape}
\usepackage{subcaption}
\usepackage{latexsym}
\usepackage{pifont}
\usepackage{multirow}

\usepackage{enumitem}
\setlist{leftmargin=6mm}
\usepackage[T1]{fontenc}
\usepackage[utf8]{inputenc}
\usepackage[font=small,labelfont=bf]{caption}

\usepackage{booktabs}

\usepackage{xcolor}
\usepackage{xspace}
\usepackage{bm}
\usepackage{amsmath}
\usepackage{amssymb}
\usepackage{placeins}
\usepackage{mdframed}   
\usepackage{listings}
\usepackage{balance}
\lstdefinestyle{promptstyle}{
    basicstyle=\normalfont\small,
    breaklines=true,
    breakatwhitespace=true,
    breakindent=0pt,
    breakautoindent=false,
    columns=flexible,
    keepspaces=true,
    showstringspaces=false,
    frame=none
}

\definecolor{CustomBlue}{RGB}{20, 81, 124}  
\definecolor{CustomLightBlue}{RGB}{47, 127, 193}  
\definecolor{CustomVeryLightBlue}{RGB}{231, 239, 250}  
\definecolor{CustomGreen}{RGB}{150, 195, 125}  
\definecolor{CustomYellow}{RGB}{243, 210, 102}  
\definecolor{CustomRed}{RGB}{216, 56, 58}  
\definecolor{CustomPink}{RGB}{247, 225, 237}  
\definecolor{CustomVeryLightPurple}{RGB}{248, 243, 249}  
\definecolor{CustomPurple}{RGB}{196, 151, 178}  
\definecolor{CustomGrayBlue}{RGB}{169, 184, 198}  

\newcommand{\ours}{\textsc{MemRerank}\xspace}

\title{\ours: Preference Memory for Personalized Product Reranking}

\author{
 \textbf{Zhiyuan Peng\textsuperscript{1}},
 \textbf{Xuyang Wu\textsuperscript{2}},
 \textbf{Huaixiao Tou\textsuperscript{2}},
 \textbf{Yi Fang\textsuperscript{1}},
 \textbf{Yu Gong\thanks{Correspondence: \texttt{gy910210@gmail.com}}\textsuperscript{2}}
\\
 \textsuperscript{1}Santa Clara University,
 \textsuperscript{2}Independent Researcher
}

\begin{document}
\maketitle
\begin{abstract}
LLM-based shopping agents increasingly rely on long purchase histories for personalization, yet naively appending raw history to the prompt is often ineffective due to noise, length, and relevance mismatch. We propose \ours, a
preference-memory framework that distills a user's purchase history into a concise, query-independent representation of shopping preferences that can be precomputed once and reused across future searches. We build a benchmark that pairs
Amazon-C4 product-search queries with the user's temporally preceding purchase history from Amazon-Review-2023, and evaluate LLM reranking over fixed top-100 candidate pools retrieved from the full product corpus. \ours trains a
Qwen2.5-7B-Instruct memory extractor with GRPO, using feedback from the downstream setwise reranker together with a lightweight deterministic memory-quality regularizer. On Electronics and Beauty \& Personal Care queries
across four first-stage retrievers, \ours attains the best macro-average MRR@10 and MRR@5, as well as the best result on a held-out retriever, while using substantially fewer memory words than raw history and MR.Rec. These results show that compact,
learned preference memory can match or exceed far longer memories at a fraction of their length, providing a practical building block for personalized agentic e-commerce search.
\end{abstract}

\section{Introduction}
Recommender systems (RS) guide users through vast item collections in domains such as e-commerce, entertainment, and social media, traditionally learning from historical interactions via collaborative filtering or deep neural models. With large language models (LLMs), recommendation is shifting from static candidate lists toward \emph{agentic} behavior, where an LLM plans, remembers, and invokes tools~\cite{DBLP:journals/tmlr/SumersYN024}. Recent perspective papers
argue that the next generation of RS will be agentic: a language core supported by tools and a hierarchical memory spanning short-term context, episodic experience, and semantic user preferences~\cite{DBLP:journals/corr/abs-2507-02097,DBLP:journals/corr/abs-2503-16734}. In such systems, memory is not a cache but a structured repository that lets agents recall prior interactions and infer latent tastes even when the user does not state them.

However, naively appending entire purchase histories or conversation transcripts to the prompt strains the context window and can dilute useful preference signal. Recent agentic frameworks therefore position the LLM as a reasoning core that consults curated memory and recommender tools rather than ingesting raw history~\cite{DBLP:journals/tois/HuangLLYLX25}, though balancing controllability, external knowledge, and long-term user context remains an open challenge for
agentic recommendation~\cite{DBLP:journals/corr/abs-2503-16734}. General memory frameworks make this concrete: Mem0~\cite{DBLP:journals/corr/abs-2504-19413} replaces large chunks of raw history with compact, salient memory, and MemAgent~\cite{DBLP:journals/corr/abs-2507-02259} selectively retains important information in long contexts. These motivate \emph{preference memory}: a compact, structured representation of user preferences, capturing within- and cross-category shopping patterns, that can be consumed directly by downstream ranking models.

We study preference memory at the product reranking stage, where its value can be measured directly against candidates returned by a first-stage retriever. We propose \ours, which distills a user's purchase history into query-independent shopping-preference memory, within-category, or both within- and cross-category, that can be reused across the user's future searches. We post-train the memory extractor with GRPO~\cite{DBLP:journals/corr/abs-2402-03300}, using feedback from a downstream setwise LLM reranker. To evaluate the full pipeline, we build a benchmark that links Amazon-C4~\cite{DBLP:journals/corr/abs-2403-03952} queries to temporally preceding Amazon-Review-2023 purchase histories and rerank fixed top-100 candidate pools retrieved from the full product corpus.

Our contributions are threefold:
\begin{itemize}
\item We release a benchmark\footnote{\url{https://huggingface.co/datasets/zhiyuanpeng/amazon-c4-user-purchase-history}}, together with our code, for personalized product reranking, containing user purchase histories, product-search queries, fixed top-100 candidate pools, and positive-product labels.
\item We design a query-independent preference-memory extractor that maps long purchase histories into compact within- and cross-category preference memory, post-trained with GRPO using downstream setwise reranking feedback.
\item Across BM25 and dense retrievers, including a held-out retriever, \ours achieves the best macro-average MRR@10 and MRR@5 while using much shorter memory than raw history and MR.Rec.
\end{itemize}

\section{Related Work}

\subsection{LLM-based Rerankers}\label{sec:llm_rerankers}

LLM-based rerankers fall into three paradigms. \textit{Pointwise} models compute a relevance score for each query-document pair using generative or discriminative scoring techniques~\cite{DBLP:journals/sigir/PonteC17,DBLP:conf/sigir/ZhuangZ21a,DBLP:conf/ecir/ZhuangLZ21,DBLP:journals/corr/abs-2404-04522,DBLP:conf/emnlp/NogueiraJPL20}. \textit{Pairwise} (and setwise) approaches compare documents against each other and use sorting or sampling algorithms to aggregate wins into a final ranking~\cite{DBLP:conf/naacl/QinJHZWYSLLMWB24,DBLP:conf/ictir/GienappFHP22,DBLP:conf/icpr/MikhailiukW0YM20,DBLP:conf/sigir/ZhuangZKZ24}. \textit{Listwise} methods produce a complete ranking in a single prompt and typically rely on zero- or few-shot prompting, sliding windows or tournament-style evaluation~\cite{DBLP:journals/corr/abs-2305-02156,DBLP:journals/corr/abs-2309-15088,DBLP:conf/emnlp/0001YMWRCYR23,DBLP:conf/www/0004LZ0MYSMY25}, while recent work fine-tunes LLMs via reinforcement learning to jointly generate reasoning and ranking for complex tasks~\cite{zhang2025rearank}. We adopt a setwise reranker for efficiency~\cite{DBLP:conf/emnlp/0001WS025} and focus on improving its input via a concise preference memory rather than altering the ranker itself.

\subsection{Personalized Product Search and Reranking}

Personalized product search ranks products by jointly modeling query relevance and user preferences. Task-specific neural architectures dominate prior work: HEM~\cite{DBLP:conf/sigir/AiZBCC17} learns shared query-user-product representations; ZAM~\cite{DBLP:conf/cikm/AiHVC19} adaptively controls when to personalize; RTM~\cite{DBLP:conf/sigir/BiAC21} performs fine-grained review-based Transformer matching; and CAMI~\cite{DBLP:conf/www/LiuDZW22} models category-aware multi-interest preferences. For re-ranking, PRM~\cite{DBLP:conf/recsys/PeiZZSLSWJGOP19} applies self-attention over candidates and user-specific signals after an initial ranking, while recent LLM-based methods refine
preferences before reranking: UR4Rec~\cite{DBLP:conf/coling/ZhangZD25} retrieves candidate-oriented user preferences and HMPPS~\cite{DBLP:conf/mm/0005LXLXRL25} applies query-aware refinement for multimodal search. These methods train task-specific retrieval, matching, or reranking models. In contrast, we study preference memory as a compact, query-independent intermediate representation that decouples preference modeling from a fixed setwise LLM reranker, so our comparisons isolate memory-construction strategies under the same candidate pool.

\subsection{Memory in Agentic Systems}\label{sec:mem_in_agentic_sys}
Memory is central to LLM-based agents. Cognitive architectures organize working, episodic, and semantic memory to support long-horizon reasoning and reduce hallucinations~\cite{DBLP:journals/tmlr/SumersYN024}. General frameworks instantiate these ideas: Mem0~\cite{DBLP:journals/corr/abs-2504-19413} dynamically extracts, consolidates, and retrieves salient user information across sessions, while MemAgent~\cite{DBLP:journals/corr/abs-2507-02259} maintains a fixed-length memory updated via reinforcement learning over long contexts. For recommendation, MR.Rec~\cite{DBLP:journals/corr/abs-2510-14629} jointly optimizes memory use and recommendation reasoning with RL, Memory-R1~\cite{DBLP:journals/corr/abs-2508-19828} learns RL controllers for what to store and retrieve, MAP~\cite{DBLP:journals/corr/abs-2505-03824} retrieves relevant entries from a per-user memory of past interactions, and Memento~\cite{DBLP:journals/corr/abs-2508-16153} adapts memory without updating base weights. Other work targets long-term dialogue, where agents struggle to recall information over many turns~\cite{DBLP:conf/acl/MaharanaLTBBF24}, through reflective or timeline-based memory management~\cite{DBLP:conf/acl/0001YHH0LSCPLI025,DBLP:conf/naacl/LiYZDWC25,DBLP:conf/naacl/OngKGCKJHLY25}. Most of these systems aim to preserve coherence or support general reasoning; in
contrast, we design preference memory tailored to downstream product ranking.

\section{Methodology}

\subsection{Dataset Construction}
\label{sec:dataset_construction}

We construct our benchmark from Amazon-Review-2023 and Amazon-C4~\cite{DBLP:journals/corr/abs-2403-03952}. Amazon-Review-2023 contains product metadata, user IDs, timestamps, and reviews across 33 categories. Amazon-C4 is a product-search dataset created by rewriting long 5-star Amazon reviews into complex search queries; each query is associated with the reviewed product, which we treat as the positive product. For each Amazon-C4 query, we recover the corresponding user's historical purchases from Amazon-Review-2023 and keep only purchases that occurred before the query's target product interaction. We remove the target positive product from the history before constructing memory inputs.

Following MR.Rec~\cite{DBLP:journals/corr/abs-2510-14629}, we rewrite Amazon-C4 queries with o3-mini using the prompt in Figure~\ref{fig:query_rewrite_prompt}. This preserves the original shopping intent while removing review-specific details that are not natural search queries. Products are represented by title and description, following Amazon-C4. For evaluation, we retrieve top-100 candidates from the full Amazon-C4 product corpus with four first-stage retrievers: BM25, BLAIR-large, Qwen3-Embedding-8B, and Linq-Embed-Mistral. All reranking methods operate on the same fixed top-100 candidate pool for a given query and retriever.

For memory extraction, we group each user's previous purchases by category. Purchases from the same category as the positive product form the within-category history, while purchases from other categories form the cross-category history. We evaluate only \emph{qualified} queries that have at least one valid within-category historical purchase after excluding the positive product; if a retriever fails to retrieve the positive product in its top 100, the reranking score for that query-retriever pair is zero. Figure~\ref{fig:data_stats_instances} shows the number of benchmark queries by target category, and Figure~\ref{fig:data_stats_history} shows that cross-category histories are typically much longer than within-category histories, motivating explicit memory compression.

\subsection{\ours Framework}

\begin{figure*}[t]
    \centering
    \includegraphics[width=\textwidth]{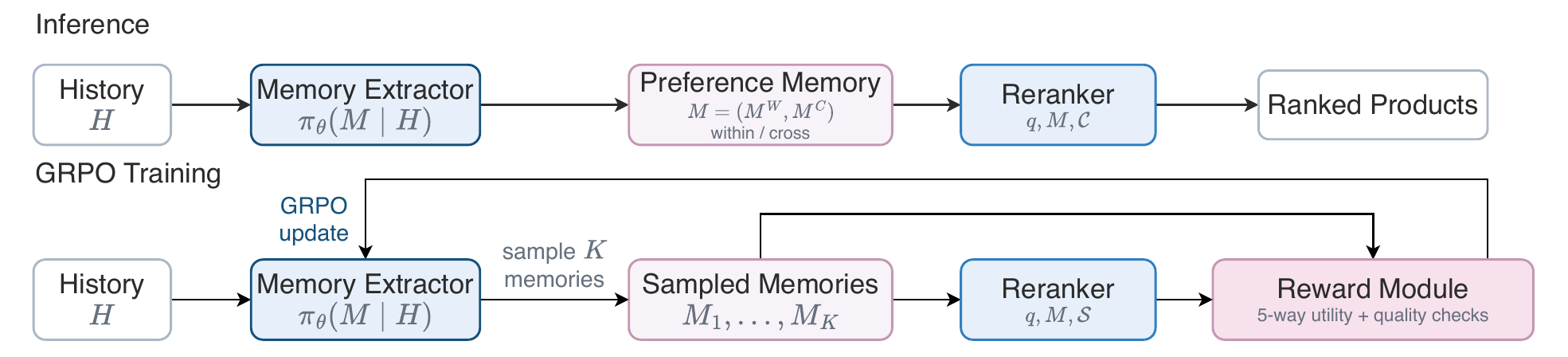}
    \caption{
    Overview of \textsc{MemRerank}. The top row shows inference, and the bottom row shows the GRPO training loop used to score sampled memories and update the memory extractor.
    }
    \label{fig:model_structure}
\end{figure*}

Figure~\ref{fig:model_structure} illustrates the two-stage \ours framework. Let $H$ denote a user's purchase history before the target interaction. The memory extractor is a policy $\pi_{\theta}(M \mid H)$ that maps $H$ into query-independent preference memory $M$. We write $M=(M^W,M^C)$, where $M^W$ summarizes within-category preferences and $M^C$ summarizes cross-category preferences. We study two memory targets: \texttt{M\_W}, which extracts only $M^W$ and leaves $M^C$ empty, and \texttt{M\_WC}, which extracts both $M^W$ and $M^C$. The \texttt{M\_W} prompt is shown in Figure~\ref{fig:M_W}; the \texttt{M\_WC} prompt is similar but includes an additional section for cross-category memory.

At inference time, the downstream reranker receives the query $q$, the extracted memory $M$, and a candidate set $\mathcal{C}$. In evaluation, $\mathcal{C}$ is the fixed top-100 product pool returned by a first-stage retriever. Instead of scoring each product independently or asking the LLM to output a full ranking at once, we use a setwise reranker that compares small candidate subsets and aggregates the decisions into a top-100 ranking. This design makes the five-candidate comparison the atomic operation for both reward collection and final setwise reranking.

During GRPO training, the extractor samples $K$ candidate memories $M_1,\ldots,M_K$. For each sampled memory $M_k$, we construct a five-candidate set $\mathcal{S}\subset\mathcal{C}$ containing one positive product $d^+$ and four retrieved negatives. The training-time reranker receives $(q,M_k,\mathcal{S})$ and is queried five times. We define the 5-way utility as
\begin{equation}\label{eq:5way_utility}
u_5(M_k) = \frac{1}{5}\sum_{j=1}^{5}\mathbf{1}[a_j = d^+],
\end{equation}
where $a_j$ is the product selected by the $j$-th reranker response. The reward module receives two signals: the 5-way utility from the reranker and deterministic quality checks computed directly on $M_k$. GRPO then updates $\pi_{\theta}$ using the relative rewards of the sampled memories. At inference time, the learned extractor produces a single memory $M$ for the user history, and repeated 5-way comparisons are aggregated into the final ranking over $\mathcal{C}$. The reported metrics are computed over the full retrieved top-100 candidate pool, not over isolated 5-way utility.

Compared with directly appending raw purchase history, \ours compresses noisy historical behavior into concise, reusable preference memory. This query-independent design is important for practical shopping agents because the same memory can be reused across future searches without reprocessing the full purchase history for every query.

\subsection{GRPO Training and Reward Design}

We train the memory extractor with GRPO using top-100-aligned 5-way reward instances. For each category, we split the data constructed in Section~\ref{sec:dataset_construction} into train, dev, and test sets with a 75:15:10 ratio. We then construct training and development instances from full-corpus top-100 retrieval results. BM25, BLAIR-large, and Qwen3-Embedding-8B are used as seen retrievers for reward-data construction, while Linq-Embed-Mistral is held out for retriever-generalization evaluation. For each query and seen retriever, we keep the pair only when the positive product appears in that retriever's top 100.

Each reward instance contains the positive product and four retrieved negatives. To expose the extractor to different parts of the top-100 distribution, we sample negatives by rank bucket: two from ranks 1-20, one from ranks 21-50, and one from ranks 51-100. We sample eight such 5-way instances per eligible query-retriever pair for training and five for development. This construction keeps the reward task cheap while making it closer to the final top-100 setwise reranking setting. 

Given an input history $H$, GRPO samples a group of memories $M_1,\ldots,M_K$ from $\pi_{\theta}(\cdot \mid H)$. For each sampled memory $M_k$, we remove evidence fields before scoring, insert the remaining memory into the downstream reranking prompt, and evaluate it on the corresponding 5-way candidate set. The reward reranker is sampled five times with the same prompt, and the utility reward is the positive-selection accuracy $u_5(M_k)$ defined in Equation~\ref{eq:5way_utility}. GRPO then updates the extractor using the relative rewards of the sampled memories, so the model is optimized for memories that make the downstream reranker more likely to select the positive product.

In addition to $u_5(M_k)$, we add a lightweight deterministic memory-quality term. The checks are intentionally category-agnostic and small relative to $u_5(M_k)$. They encourage well-formed memory fields, concise preference statements, and non-empty within-category memory, while penalizing malformed tags, repetitive or overlong output, placeholder text, and unsupported evidence-like content. Let $q(M_k)$ denote the clipped quality score and $\lambda$ its weight. The reward used by GRPO is
\begin{equation}
r(H,M_k) = u_5(M_k) + \lambda q(M_k).
\end{equation}
This design keeps downstream reranking utility as the main optimization target and uses quality checks only as regularization. It avoids requiring gold preference-memory annotations, and it trains the extractor for its actual role in the system: producing compact reusable memory that improves personalized product reranking.

\begin{table*}[t]
\centering
\scriptsize
\setlength{\tabcolsep}{2.0pt}
\renewcommand{\arraystretch}{1.08}
\resizebox{\textwidth}{!}{
\begin{tabular}{@{}clcccccccccc@{}}
\toprule
\multirow{2}{*}[-2.0ex]{\textbf{Category}} &
\multicolumn{1}{c}{\multirow{2}{*}[-2.0ex]{\textbf{Retriever}}} &
\multirow{2}{*}[-2.0ex]{\textbf{R@100}} &
\multirow{2}{*}[-2.0ex]{\shortstack{\textbf{Ret.}\\\textbf{MRR@10}}} &
\multicolumn{7}{c}{\textbf{Setwise reranking with o4-mini, MRR@10}} &
\multirow{2}{*}[-2.0ex]{$\boldsymbol{\Delta_{\mathrm{best}}}$} \\
\cmidrule(lr){5-11}
& & & &
\textbf{No Mem.} &
\textbf{Raw Hist.} &
\textbf{GPT-5.5} &
\textbf{MR.Rec} &
\textbf{Mem0} &
\shortstack{\textbf{\textsc{MemRerank}}\\\textbf{w/o RL}} &
\textbf{\textsc{MemRerank}} &
\\
\midrule

\multirow{5}{*}{\textbf{Electronics}}
& BM25                          & 16.78 & 1.44 & 7.15 & 6.15 & 5.78 & 6.27 & \underline{7.92} & 6.77 & \textbf{7.93} & +0.01 \\
& BLAIR-large                   & 41.96 & 3.15 & 9.78 & 10.85 & 10.54 & \textbf{11.81} & \underline{11.36} & 10.13 & 11.28 & -0.53 \\
& Qwen3-Embedding-8B            & 39.16 & 5.91 & 8.25 & 11.11 & 10.61 & \underline{11.19} & 11.01 & 10.74 & \textbf{11.44} & +0.25 \\
& Linq-Embed-Mistral$^\dagger$  & 41.26 & 3.80 & 11.15 & 12.08 & 12.70 & \underline{13.07} & 11.10 & 11.46 & \textbf{14.36} & +1.29 \\
\cmidrule(lr){2-12}
& All Avg.                      & 34.79 & 3.58 & 9.08 & 10.05 & 9.91 & \underline{10.59} & 10.35 & 9.78 & \textbf{11.25} & +0.66 \\
\midrule

\multirow{5}{*}{\textbf{Beauty \& Personal Care}}
& BM25                          & 13.16 & 1.13 & 1.56 & 1.99 & 1.64 & 1.26 & 1.66 & \underline{2.01} & \textbf{2.16} & +0.15 \\
& BLAIR-large                   & 27.19 & 4.86 & 3.70 & 4.81 & 3.96 & \textbf{5.63} & \underline{5.03} & 4.14 & 4.11 & -1.52 \\
& Qwen3-Embedding-8B            & 39.47 & 3.64 & \textbf{6.81} & 5.15 & 5.31 & 4.74 & 4.63 & 5.12 & \underline{5.94} & -0.87 \\
& Linq-Embed-Mistral$^\dagger$  & 37.72 & 3.77 & 5.26 & 5.06 & 5.20 & \underline{5.30} & 3.78 & 5.00 & \textbf{6.77} & +1.47 \\
\cmidrule(lr){2-12}
& All Avg.                      & 29.39 & 3.35 & \underline{4.33} & 4.25 & 4.03 & 4.23 & 3.78 & 4.07 & \textbf{4.75} & +0.42 \\
\midrule

\multicolumn{2}{@{}l}{\textbf{Macro Avg.}}
& 32.09 & 3.47 & 6.71 & 7.15 & 6.97 & \underline{7.41} & 7.07 & 6.93 & \textbf{8.00} & +0.59 \\

\bottomrule
\end{tabular}
}
\caption{
Top-100 personalized product reranking on test queries, reported as MRR@10 ($\times 100$).
R@100 and Ret. MRR@10 are first-stage retrieval metrics; the remaining columns are o4-mini setwise reranking results using the same candidate pool within each row.
Linq-Embed-Mistral$^\dagger$ is held out from training-negative construction.
Bold and underline mark the best and second-best reranking methods; $\Delta_{\mathrm{best}}$ is the improvement of \textsc{MemRerank} over the strongest baseline.
}
\label{tab:top100_main}
\end{table*}

\section{Experimental Setup}
\subsection{Baselines}

We compare our method against several types of baselines that differ in how user preference information is provided to the reranker.

\paragraph{No memory.}
This setting performs direct product reranking without explicit user memory. The reranker receives only the query and candidate products, making it the reference point for measuring the value of preference memory.

\paragraph{Raw history.}
This baseline appends raw purchase-history products directly to the reranking prompt instead of extracting memory. It tests whether explicit memory compression is necessary, or whether the reranker can use the recent history products by itself.

\paragraph{LLM-extracted memory baselines.}
We compare against two direct memory-extraction baselines.
\emph{GPT-5.5} uses gpt-5.5 to extract preference memory with the same memory target as the evaluated setting.
\textsc{MemRerank} w/o RL uses the same Qwen2.5-7B-Instruct backbone and extraction prompt as \ours, but without GRPO post-training.

\paragraph{External memory baselines.}
We further compare with two memory-centric systems, \emph{MR.Rec} and \emph{Mem0}. To keep the comparison controlled, we run both methods with Qwen2.5-7B-Instruct and the same metadata-only history window used by \ours. MR.Rec first summarizes category-level preferences and then constructs a user profile. Mem0 extracts atomic preference facts from the user's history. We render their outputs into the same memory fields consumed by the downstream reranker.

\paragraph{Our method.}
\ours uses our RL-trained memory extractor to produce query-independent preference memory optimized for downstream setwise reranking.

\subsection{Implementation Details}

We use Qwen2.5-7B-Instruct as the memory-extractor backbone. For each category, we select the memory target and checkpoint using development-set reranking performance. The main table uses within+cross memory for Electronics and within-category memory for Beauty \& Personal Care. We train for five epochs with rollout size 8 and use o4-mini as the reward reranker. The reward reranker is sampled five times per training instance; the utility reward is the fraction of responses that select the positive product.

For the final top-100 setwise evaluation, all methods use o4-mini with temperature 1.0, top\_p 1.0, and reasoning\_effort=low. We cap candidate-product metadata and history-product metadata at 120 words. Evidence fields produced by memory extractors are removed before both reward scoring and final reranking; the evidence-containing memories are retained only for auditing. We report MRR@10 as the primary metric and place MRR@5 in Appendix~\ref{tab:top100_mrr5_appendix}.

\begin{figure}[t]
    \centering
    \begin{subfigure}{0.95\columnwidth}
        \centering
        \includegraphics[width=\linewidth]{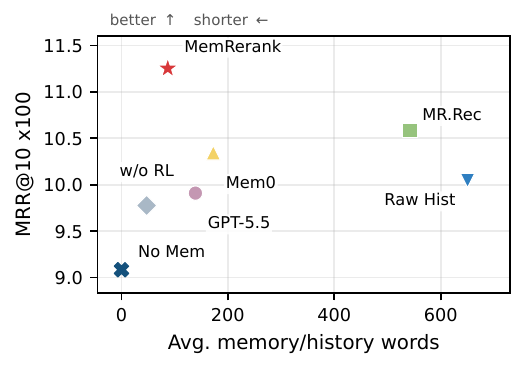}
        \caption{Electronics}
        \label{fig:memory_words_mrr10_electronics}
    \end{subfigure}

    \vspace{0.4em}

    \begin{subfigure}{0.95\columnwidth}
        \centering
        \includegraphics[width=\linewidth]{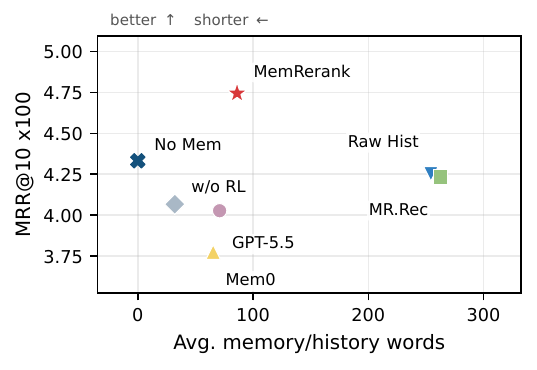}
        \caption{Beauty \& Personal Care}
        \label{fig:memory_words_mrr10_beauty}
    \end{subfigure}
    \caption{
    Effectiveness-memory tradeoff on Electronics and Beauty \& Personal Care.
    }
    \label{fig:memory_words_mrr10}
\end{figure}

\section{Experimental Results and Analysis}

\subsection{Main Results}

Table~\ref{tab:top100_main} shows that \ours achieves the strongest macro-average top-100 reranking performance. On Electronics, \ours improves the category average MRR@10 from 10.59 for the strongest baseline to 11.25. On Beauty \& Personal Care, \ours improves the category average from 4.33 to 4.75. Across the two categories, \ours improves the macro average from 7.41 to 8.00. The gains are not uniform for every retriever row, but they are consistent at the category-average level, which is the main metric for evaluating robustness across retrieval systems.

The held-out Linq-Embed-Mistral rows test retriever generalization. Linq is not used when constructing training negatives or selecting checkpoints. \ours obtains the best Linq MRR@10 in both categories, improving Electronics from 13.07 to 14.36 and Beauty \& Personal Care from 5.30 to 6.77 over the strongest baseline in each row. This suggests that the learned memory extractor is not only fitting the failure patterns of the seen retrievers.

Figure~\ref{fig:memory_words_mrr10} further shows that the improvement does not come from simply increasing prompt length. Raw history and MR.Rec use much longer history or memory text, yet \ours achieves higher MRR@10 with a compact preference memory. This supports the central design choice of learning a concise intermediate memory instead of exposing the reranker to the full purchase history.

\subsection{Ablation Study}

Table~\ref{tab:ablation_target_quality} studies two design choices:
whether the memory extractor uses only within-category history or both within- and cross-category history, and whether the reward includes the deterministic memory-quality regularizer.
All variants use the same setting as Table~\ref{tab:top100_main}.
The results show that the best memory target is category-dependent: Electronics benefits more from within+cross memory, while Beauty \& Personal Care is better served by within-category memory.
The quality regularizer improves the selected setting in both categories.

\begin{table}[t]
\centering
\scriptsize
\setlength{\tabcolsep}{2.5pt}
\renewcommand{\arraystretch}{1.05}
\resizebox{\columnwidth}{!}{
\begin{tabular}{@{}llccc@{}}
\toprule
\textbf{Category} &
\textbf{Memory} &
\textbf{$\lambda=0.0$} &
\textbf{$\lambda=0.4$} &
\textbf{$\Delta$} \\
\midrule
\multirow{2}{*}{Electronics}
& \texttt{M\_W} & 9.98 & 10.71 & +0.73 \\
& \texttt{M\_WC} & 10.36 & \textbf{11.25} & +0.89 \\
\midrule
\multirow{2}{*}{Beauty \& Personal Care}
& \texttt{M\_W} & 3.35 & \textbf{4.75} & +1.40 \\
& \texttt{M\_WC} & 3.23 & 3.74 & +0.51 \\
\bottomrule
\end{tabular}
}
\caption{
Ablation of memory target and quality reward.
Values are average MRR@10 over the four retrievers, multiplied by 100.
$\lambda$ is the quality-reward weight.
$\Delta$ is $\lambda=0.4$ minus $\lambda=0.0$.
}
\label{tab:ablation_target_quality}
\end{table}

\paragraph{Memory Target}
The best memory target is category-dependent. Electronics benefits from including cross-category history: \texttt{M\_WC} reaches 11.25 MRR@10, compared with 10.71 for \texttt{M\_W}. This is plausible because electronics purchases often share transferable preferences such as brand ecosystem, durability, compatibility, or budget. In contrast, Beauty \& Personal Care performs better with within-category memory only: \texttt{M\_W} reaches 4.75, while \texttt{M\_WC} reaches 3.74. For this category, unrelated cross-category purchases appear more likely to introduce noise than useful preference signal.

\paragraph{Memory Quality Reward}
The quality reward improves average MRR@10 in all four category-target comparisons. The gains are +0.73 and +0.89 on Electronics, and +1.40 and +0.51 on Beauty \& Personal Care. Because the quality term is deterministic and much smaller than the reranking utility term, these improvements suggest that simple constraints on memory structure, conciseness, and support can regularize RL training without replacing task-level feedback.

\section{Conclusion}

We present \ours, a framework for personalized product reranking with compact, query-independent preference memory. Instead of directly feeding long purchase histories to an LLM reranker, \ours trains a Qwen2.5 memory extractor with downstream setwise reranking feedback and lightweight memory-quality regularization. On top-100 reranking over Electronics and Beauty \& Personal Care, \ours achieves the best macro-average MRR@10 across BM25 and dense retrievers, including a held-out dense retriever not used for training-negative construction. The results suggest that learned preference memory can turn historical shopping behavior into concise and reusable personalization signals for agentic e-commerce search.

\section*{Limitations}

This work focuses on textual product metadata and purchase histories from two Amazon-C4 categories. Broader evaluation across more categories, multilingual queries, and multimodal product signals such as images would further test the generality of the approach. Both the training reward and the final evaluation rely on the proprietary o4-mini API. This reflects a realistic agentic-search setting, but API-based reranking is more expensive than self-hosted open-weight reranking at large scale. Future work can explore lower-cost distilled rerankers or hybrid evaluation protocols while keeping the same preference-memory interface.


\bibliography{custom}

\newpage
\appendix
\onecolumn

\section{Appendix}

\subsection{MRR@5}\label{tab:top100_mrr5_appendix}

\begin{table*}[h]
\centering
\scriptsize
\setlength{\tabcolsep}{2.0pt}
\renewcommand{\arraystretch}{1.08}
\resizebox{\textwidth}{!}{
\begin{tabular}{@{}clcccccccccc@{}}
\toprule
\multirow{2}{*}[-2.0ex]{\textbf{Category}} &
\multicolumn{1}{c}{\multirow{2}{*}[-2.0ex]{\textbf{Retriever}}} &
\multirow{2}{*}[-2.0ex]{\textbf{R@100}} &
\multirow{2}{*}[-2.0ex]{\shortstack{\textbf{Ret.}\\\textbf{MRR@5}}} &
\multicolumn{7}{c}{\textbf{Setwise reranking with o4-mini, MRR@5}} &
\multirow{2}{*}[-2.0ex]{$\boldsymbol{\Delta_{\mathrm{best}}}$} \\
\cmidrule(lr){5-11}
& & & &
\textbf{No Mem.} &
\textbf{Raw Hist.} &
\textbf{GPT-5.5} &
\textbf{MR.Rec} &
\textbf{Mem0} &
\shortstack{\textbf{\textsc{MemRerank}}\\\textbf{w/o RL}} &
\textbf{\textsc{MemRerank}} &
\\
\midrule

\multirow{5}{*}{\textbf{Electronics}}
& BM25                          & 16.78 & 1.04 & 6.55 & 5.61 & 5.09 & 5.62 & \textbf{7.67} & 6.66 & \underline{7.53} & -0.14 \\
& BLAIR-large                   & 41.96 & 2.16 & 8.75 & 9.92 & 9.53 & \textbf{10.80} & \underline{10.44} & 9.13 & 10.38 & -0.42 \\
& Qwen3-Embedding-8B            & 39.16 & 5.48 & 7.11 & \textbf{10.40} & 9.32 & \underline{10.34} & 10.19 & 9.93 & 10.30 & -0.10 \\
& Linq-Embed-Mistral$^\dagger$  & 41.26 & 3.14 & 10.56 & 11.18 & 11.86 & \underline{12.17} & 9.99 & 10.51 & \textbf{13.33} & +1.16 \\
\cmidrule(lr){2-12}
& All Avg.                      & 34.79 & 2.96 & 8.24 & 9.27 & 8.95 & \underline{9.73} & 9.57 & 9.06 & \textbf{10.39} & +0.66 \\
\midrule

\multirow{5}{*}{\textbf{Beauty \& Personal Care}}
& BM25                          & 13.16 & 0.88 & 1.10 & \underline{1.90} & 1.39 & 0.88 & 1.54 & 1.75 & \textbf{2.05} & +0.15 \\
& BLAIR-large                   & 27.19 & 4.46 & 3.29 & 4.33 & 3.61 & \textbf{5.03} & \underline{4.56} & 3.54 & 3.49 & -1.54 \\
& Qwen3-Embedding-8B            & 39.47 & 2.95 & \textbf{6.51} & 5.00 & 4.53 & 4.08 & 3.96 & 4.90 & \underline{5.73} & -0.78 \\
& Linq-Embed-Mistral$^\dagger$  & 37.72 & 3.39 & 4.91 & 4.71 & \underline{5.10} & 5.07 & 3.46 & 4.85 & \textbf{6.32} & +1.22 \\
\cmidrule(lr){2-12}
& All Avg.                      & 29.39 & 2.92 & 3.95 & \underline{3.98} & 3.66 & 3.77 & 3.38 & 3.76 & \textbf{4.40} & +0.42 \\
\midrule

\multicolumn{2}{@{}l}{\textbf{Macro Avg.}}
& 32.09 & 2.94 & 6.10 & 6.63 & 6.31 & \underline{6.75} & 6.48 & 6.41 & \textbf{7.40} & +0.65 \\

\bottomrule
\end{tabular}
}
\caption{
MRR@5 results for the same test queries as Table~\ref{tab:top100_main}.
}
\end{table*}

\subsection{Prompts}

\begin{figure}[h]
    \centering
    \begin{mdframed}[
        linecolor=black!60,
        linewidth=1pt,
        roundcorner=10pt,
        backgroundcolor=gray!5,
        shadow=true,
        shadowsize=5pt,
        shadowcolor=black!40,
        skipabove=10pt,
        skipbelow=10pt,
        innertopmargin=0pt,
        innerbottommargin=0pt,
        innerleftmargin=10pt,
        innerrightmargin=10pt
    ]
    \begin{lstlisting}[style=promptstyle]
      You are a memory extraction system for personalized recommendation.

      You will receive individual purchase items from a user's shopping history, one at a time.
      Each item may include:
      - Category information (e.g., [Electronics])
      - Product metadata (title, description, features)
      - Optional user review text
    
      Your task:
      Extract the most important shopping preferences or decision drivers from each purchase item that may help future purchase decisions.
      
      Rules:
      - Only extract what is directly supported by the input
      - Create atomic, generalizable statements useful for future recommendations
      - Avoid copying long descriptions; abstract into preference facts
      - Skip product IDs, order details, timestamps
      - Update or consolidate memories as you see patterns across multiple items
      - Return at most 3 most important preference facts
      - If nothing meaningful can be extracted, return empty list
    
      Output format:
      You MUST return ONLY raw JSON without any markdown formatting, code blocks, or extra text.
      Format: {"facts": ["preference 1", "preference 2", ...]}
      
      Do NOT wrap the JSON in ```json or ``` markers.
      Do NOT add any explanation before or after the JSON.
      
      Example valid output:
      {"facts": ["User prefers wireless headphones", "User values noise cancellation features"]}
    \end{lstlisting}
    \end{mdframed}
    \caption{Mem0 prompt}
    \label{fig:mem0_prompt}
\end{figure}

\begin{figure}[h]
    \centering
    \begin{mdframed}[
        linecolor=black!60,
        linewidth=1pt,
        roundcorner=10pt,
        backgroundcolor=gray!5,
        shadow=true,
        shadowsize=5pt,
        shadowcolor=black!40,
        skipabove=10pt,
        skipbelow=10pt,
        innertopmargin=0pt,
        innerbottommargin=0pt,
        innerleftmargin=10pt,
        innerrightmargin=10pt
    ]
    \begin{lstlisting}[style=promptstyle]
      You are given the following information about a user's purchase history in a specific category:

      - Category: {category}
      - User's purchased items in this category, each with its metadata and the user's review:
        {meta_reviews}
    
      Your task:
      Analyze the items (including their metadata and the user's reviews) and summarize the user's preferences in this category.
      The summary should capture consistent patterns across items and reviews (e.g., favored brands, preferred price range, styles, features, quality expectations).
      Be as detailed as possible, but do not fabricate information that is not supported by the input.
    
      Directly output the preference summary (string) below:
    \end{lstlisting}
    \end{mdframed}
    \caption{MR.Rec Preference Patterns Extraction Prompt}
    \label{fig:mr.rec_pp_prompt}
\end{figure}

\begin{figure}[h]
    \centering
    \begin{mdframed}[
        linecolor=black!60,
        linewidth=1pt,
        roundcorner=10pt,
        backgroundcolor=gray!5,
        shadow=true,
        shadowsize=5pt,
        shadowcolor=black!40,
        skipabove=10pt,
        skipbelow=10pt,
        innertopmargin=0pt,
        innerbottommargin=0pt,
        innerleftmargin=10pt,
        innerrightmargin=10pt
    ]
    \begin{lstlisting}[style=promptstyle]
      You are given the following information about a user's preferences across different categories or aspects:
      {preference_patterns}
    
      Your task:
      Based on these preferences, infer the user's overall profile.
      The profile should summarize general traits, patterns, and tendencies that can be reasonably inferred from the given preferences
      (e.g., spending habits, brand inclination, style choices, feature priorities, quality expectations, lifestyle hints).
      Do not fabricate information that is not supported by the input.
    
      Directly output the profile (string) below:
    \end{lstlisting}
    \end{mdframed}
    \caption{MR.Rec User Profile Extraction Prompt}
    \label{fig:mr.rec_up_prompt}
\end{figure}

\begin{figure}[h]
    \centering
    \begin{mdframed}[
        linecolor=black!60,
        linewidth=1pt,
        roundcorner=10pt,
        backgroundcolor=gray!5,
        shadow=true,
        shadowsize=5pt,
        shadowcolor=black!40,
        skipabove=10pt,
        skipbelow=10pt,
        innertopmargin=0pt,
        innerbottommargin=0pt,
        innerleftmargin=10pt,
        innerrightmargin=10pt
    ]
    \begin{lstlisting}[style=promptstyle]
    system: 
    Rewrite the "Original Query" into a single, casual, conversational query. Rules:
    - Keep: 1 core item + 1 core use case (e.g., "a laptop for gaming").
    - Delete: All other details (including brand, specs, price, personal preferences, etc.).
    - Tone: Natural and conversational.
    - Format: Must be a single sentence.
    user: 
    Original Query: {query}
    Rewritten Query:
    \end{lstlisting}
    \end{mdframed}
    \caption{Query Rewrite Prompt}
    \label{fig:query_rewrite_prompt}
\end{figure}

\begin{figure}[h]
    \centering
    \begin{mdframed}[
        linecolor=black!60,
        linewidth=1pt,
        roundcorner=10pt,
        backgroundcolor=gray!5,
        shadow=true,
        shadowsize=5pt,
        shadowcolor=black!40,
        skipabove=10pt,
        skipbelow=10pt,
        innertopmargin=0pt,
        innerbottommargin=0pt,
        innerleftmargin=10pt,
        innerrightmargin=10pt
    ]
    \begin{lstlisting}[style=promptstyle]
    system:
    You are an intelligent shopping assistant specialized in selecting the most relevant product from a list of candidates based on a customer's search query.
    user:
    Given a customer search query "{query}", which of the following products is the most relevant to the query?

    Below is user preference memory that has been precomputed from the user's historical behavior.
    Some memory blocks may be empty or omitted; if a block is not present, you should simply ignore it.

    WITHIN-CATEGORY PREFERENCE MEMORY (optional, may be based on M, R, or M+R)
    Category: {category_name}
    ------------------------------
    {within_memory}

    CROSS-CATEGORY PREFERENCE MEMORY (optional, may be based on M, R, or M+R)
    ------------------------------
    {cross_memory}

    Candidate products:
    {passages}

    When choosing the most relevant product:
    - First, make sure the product matches the customer's query.
    - If preference memory is provided:
      * Use the within-category memory (if present) to judge how well each candidate aligns with the user's stable preferences for this type of product.
      * Use the cross-category memory (if present) to ensure the choice fits the user's general shopping style (e.g., budget, quality expectations, brand preferences) and to break ties.
    - However, the query relevance must remain the primary criterion.

    Output ONLY the product label (e.g., A, B, C) of the most relevant product wrapped in <answer> tags, like <answer>A</answer>. Do not output any other text.
    \end{lstlisting}
    \end{mdframed}
    \caption{Reranking Prompt with Memory}
    \label{fig:v1_mem}
\end{figure}

\begin{figure}[h]
    \centering
    \begin{mdframed}[
        linecolor=black!60,
        linewidth=1pt,
        roundcorner=10pt,
        backgroundcolor=gray!5,
        shadow=true,
        shadowsize=5pt,
        shadowcolor=black!40,
        skipabove=10pt,
        skipbelow=10pt,
        innertopmargin=0pt,
        innerbottommargin=0pt,
        innerleftmargin=10pt,
        innerrightmargin=10pt
    ]
    \begin{lstlisting}[style=promptstyle]
    system: 
    You are a user preference memory extractor for an e-commerce recommendation system.
    Your goal is to summarize a user's stable preferences for ONE product category
    based ONLY on metadata (M) of their past purchases in that category.

    Definition:
    - M (meta-data) = product TITLE + product DESCRIPTION.

    The summary must be independent of any specific search query, so it can be reused
    for future tasks for the same user.

    user: 
    Below is this user's purchase history in category "{category_name}".

    Each entry contains:
    - [M]: product metadata (TITLE + DESCRIPTION)

    ==============================
    WITHIN-CATEGORY HISTORY (META ONLY)
    Category: {category_name}
    ------------------------------
    {within_M_history}
    ==============================

    Using ONLY the provided metadata (M) for these past purchases:
    - Infer the user's stable preferences specifically for category "{category_name}".
    - Discover 3-6 aspects that are most strongly supported by the purchase history and most useful for distinguishing this user's preferences in this category.
    - The aspect names are NOT fixed. You may include common aspects (e.g., typical brands, feature priorities, style/aesthetics, quality level, price range or budget level, common usage scenarios), but these are examples-not a required checklist. Do NOT output an aspect unless it is clearly evidenced in the history.
    - Avoid generic headings like "features" or "style" unless you specify what features/style (e.g., "noise cancelling", "USB-C", "wireless", "minimalist design").
    - Each bullet MUST include an Evidence field:
        * Evidence must be copied verbatim from the provided purchase history metadata (TITLE/DESCRIPTION fragments).
        * Provide 1-2 short evidence snippets, each $\leq$ 10 words.
        * Format Evidence as a JSON-like list of quoted snippets: ["snippet1"] or ["snippet1", "snippet2"].
    - Make each Preference actionable for ranking products in "{category_name}" (state what to prefer/avoid).
    - Keep the memory concise:
        * Output 3-6 bullets total (output fewer if evidence is weak; do NOT pad).
        * Each bullet must be a single line.
        * Each Preference should be $\leq$ 20 words.
    - Do NOT repeat the same aspect name across bullets. Use short aspect labels (2-4 words) and avoid near-duplicates (e.g., "Brand" vs "Brands").
    - If the history does not strongly support an aspect, omit it rather than guessing.
    - If evidenced, you may also include aspects like compatibility/ecosystem, portability/travel, durability/warranty, or bundle/kit preferences.

    The memory must be query-independent: describe what the user generally likes
    or dislikes in this category, not what they might want in a single session.

    Output your summary in the following format:

    <within_memory>
    - Aspect: <aspect_1>; Preference: <user's preference for this aspect>; Evidence: ["<snippet1>", "<snippet2>"]
    - Aspect: <aspect_2>; Preference: <user's preference for this aspect>; Evidence: ["<snippet1>"]
    ...
    </within_memory>

    Do NOT output anything outside the <within_memory> tags.
    \end{lstlisting}
    \end{mdframed}
    \caption{Within-category Shopping Preference Memory Extraction Prompt (M\_W)}
    \label{fig:M_W}
\end{figure}

\clearpage
\subsection{Dataset}

\begin{figure}[H]
    \centering
    \begin{subfigure}{0.86\textwidth}
        \centering
        \includegraphics[width=\textwidth]{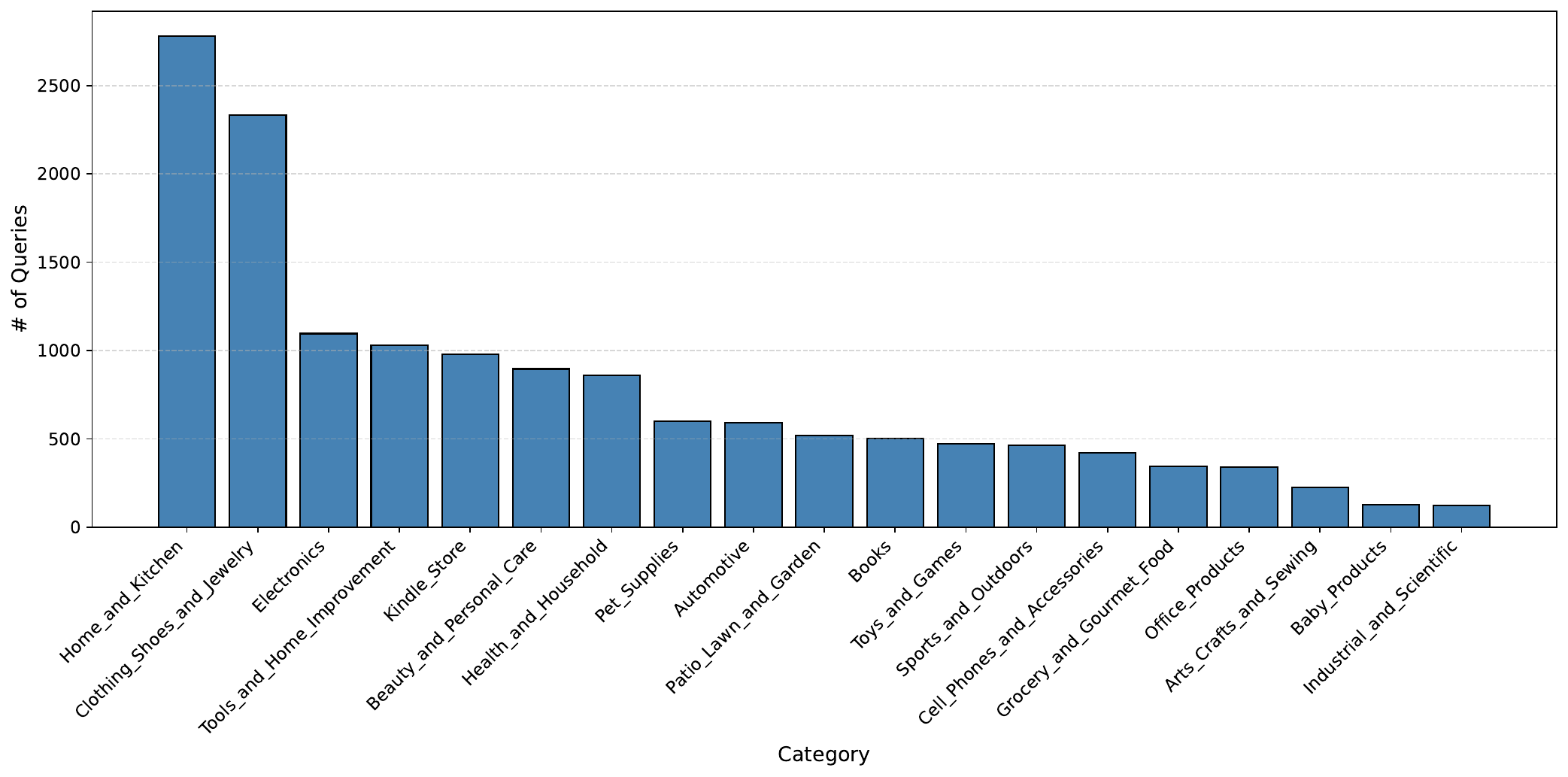}
        \caption{Number of benchmark queries per target category. Categories with fewer than 100 queries are omitted for readability.}
        \label{fig:data_stats_instances}
    \end{subfigure}

    \vspace{0.3em}

    \begin{subfigure}{0.86\textwidth}
        \centering
        \includegraphics[width=\textwidth]{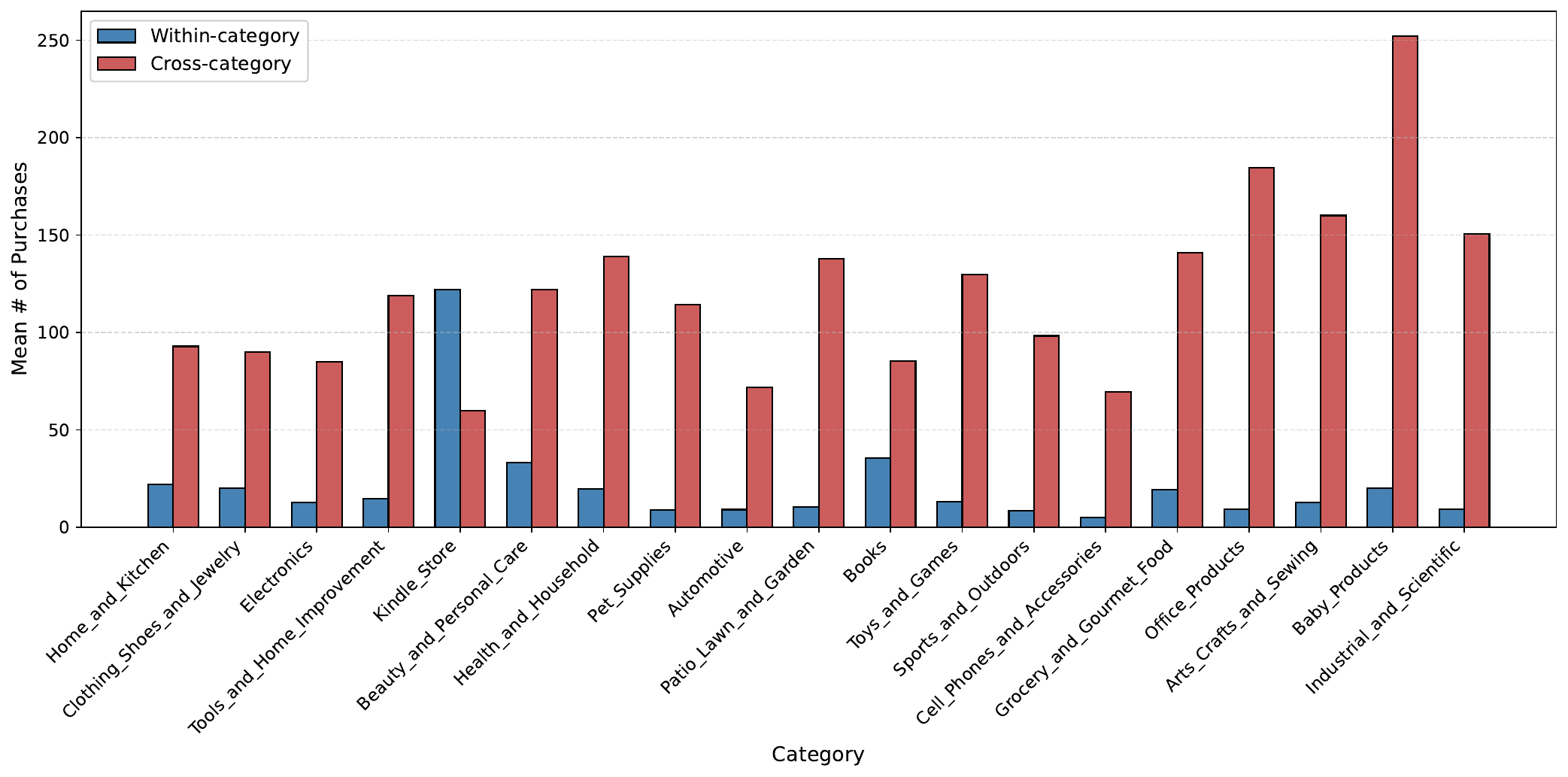}
        \caption{Average numbers of within-category and cross-category purchases per query.}
        \label{fig:data_stats_history}
    \end{subfigure}

    \caption{Dataset statistics. Categories are sorted by the number of benchmark queries, and categories with fewer than 100 queries are omitted for readability. The top figure shows the number of benchmark queries for each target category. The bottom figure compares the average lengths of within-category and cross-category purchase histories after excluding the target positive product. Cross-category histories are generally longer, motivating separate modeling of within-category and cross-category preference memory.}
    \label{fig:data_stats}
\end{figure}

\end{document}